\documentclass[preprint,12pt,authoryear]{elsarticle}
\usepackage{natbib}  
\usepackage{algorithmic}
\usepackage{diagbox}  
\usepackage{pifont}
\usepackage{amsmath,graphicx,hyperref,amsfonts}
\newcommand{\cxmark}{\ding{55}}
\usepackage{threeparttable} 
\usepackage{xcolor}
\usepackage{amssymb}
\usepackage{mathrsfs}
\usepackage{multirow}
\usepackage[T1]{fontenc}
\usepackage{pdfpages}
\usepackage{subfigure}

\journal{Medical Image Analysis}

\begin{document}
%\includepdf[pages=1-2]{Titlepage.pdf}
\begin{frontmatter}

\title{Text-Conditioned Multi-Expert Regression Framework for Fully Automated Multi-Abutment Design}

 \author[label1,label2,label3]{Mianjie Zheng\fnmark[1]}
%\author[label1,label2,label3]{Mianjie Zheng}
\ead{2400101029@mails.szu.edu.cn}
 \author[label1,label2,label3]{Xinquan Yang\fnmark[1]}
%\author[label1,label2,label3]{Xinquan Yang\corref{mycorrespondingauthor}}
\ead{yangxinquan2021@email.szu.edu.cn}
\author[label1,label2,label3]{Xuefen Liu}
\ead{2400101034@mails.szu.edu.cn}
\author[label5]{Xuguang Li}
\ead{lixuguang@szu.edu.cn}
\author[label1,label2,label3]{Kun Tang}
\ead{2500101065@mails.szu.edu.cn}
\author[label5]{He Meng\corref{mycorrespondingauthor}}
\ead{menghe@szu.edu.cn}
\author[label1,label2,label3,label4]{Linlin~Shen\corref{mycorrespondingauthor}}
\ead{llshen@szu.edu.cn}
\cortext[mycorrespondingauthor]{Corresponding author}

\address[label1]{College of Computer Science and Software Engineering, Shenzhen University, Shenzhen, China}
\address[label2]{AI Research Center for Medical Image Analysis and Diagnosis, Shenzhen University, Shenzhen, China}
\address[label3]{National Engineering Laboratory for Big Data System Computing Technology, Shenzhen University, China}
\address[label4]{School of Artificial Intelligence, Shenzhen University, Shenzhen, China}
\address[label5]{Department of Stomatology, Shenzhen University General Hospital, Shenzhen, China}

 \fntext[1]{M. Zheng and X. Yang contributed equally to this work and are co-first authors.}
\begin{abstract}
% Implant abutment design plays a critical role in dental rehabilitation, where geometric precision is essential to reduce biological and mechanical complications. Existing workflows are largely manual or semi-automated, requiring intensive clinician intervention and lacking scalability for multi-abutment scenarios. We propose TEMAD, a fully automated text-conditioned multi-expert architecture for multi-abutment design, integrating edentulous site localization and implant system–compatible abutment parameter regression into a unified framework. An Implant Area Prediction Module (IAPM) is introduced to automatically localize implant sites, eliminating manual annotation. A pre-trained mesh encoder extracts geometric features from intraoral scans, and a Tooth-Conditioned Feature-wise Linear Modulation (TC-FiLM) module adaptively calibrates mesh representations using tooth embeddings to enable position-specific feature modulation. Furthermore, a System-Prompted Mixture-of-Experts (SPMoE) mechanism leverages implant system prompts to guide expert selection, enabling system-aware regression. Extensive experiments on multi-abutment datasets demonstrate that TEMAD achieves improved geometric accuracy and implant system compatibility, particularly in multi-abutment settings, validating its effectiveness for fully automated dental implant planning.
Dental implant abutments serve as the geometric and biomechanical interface between the implant fixture and the prosthetic crown, yet their design relies heavily on manual effort and is time-consuming. Although deep neural networks have been proposed to assist dentists in designing abutments, most existing approaches remain largely manual or semi-automated, requiring substantial clinician intervention and lacking scalability in multi-abutment scenarios.
To address these limitations, we propose TEMAD—a fully automated, text-conditioned multi-expert architecture for multi-abutment design. This framework integrates implant site localization and implant system–compatible abutment parameter regression into a unified pipeline. 
Specifically, we introduce an Implant Site Identification Network (ISIN) to automatically localize implant sites and provide this information to the subsequent multi-abutment regression network. We further design a Tooth-Conditioned Feature-wise Linear Modulation (TC-FiLM) module, which adaptively calibrates mesh representations using tooth embeddings to enable position-specific feature modulation. Additionally, a System-Prompted Mixture-of-Experts (SPMoE) mechanism leverages implant system prompts to guide expert selection, ensuring system-aware regression.
Extensive experiments on a large-scale abutment design dataset show that TEMAD achieves state-of-the-art performance compared to existing methods, particularly in multi-abutment settings, validating its effectiveness for fully automated dental implant planning.
\end{abstract}

\end{frontmatter}

\section{Introduction}
\begin{figure*}\centering
\includegraphics[scale=0.35]{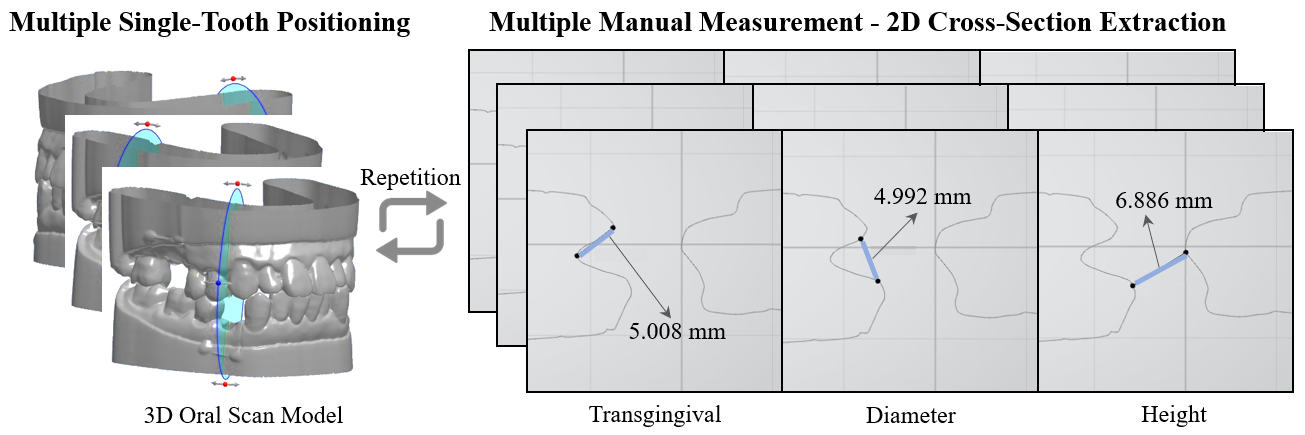}
\caption{Illustration of the traditional CAD–CAM workflow for multi-abutment design, where each implant site must be positioned and measured individually before designing the corresponding abutment based on the measured parameters.}
\label{manual_measurement}
\end{figure*}

Dental implant abutment is a critical component in implant‑supported prosthetic rehabilitation, serving as the geometric and biomechanical interface between the implant fixture and the prosthetic crown. Its structural design directly influences mechanical stability and long‑term occlusal function~\cite{caricasulo2018influence}.
In current clinical practice, abutment design relies on manual measurements taken from physical dental models or digital intraoral scans~\cite{tartea2023comparative, benakatti2021dental}. Key parameters include transgingival height, implant diameter, and gingivo‑marginal distance. As shown in Fig.~\ref{manual_measurement}, even with CAD‑CAM‑assisted digital models, clinicians must manually locate each implant site and extract cross‑sectional measurements to determine the restoration‑space parameters. Each abutment requires expert evaluation and iterative refinement, making the process time‑consuming and labor‑intensive~\cite{shah2023literature}.
When multiple abutments are needed within the same intraoral scan, these steps must be repeated for every implant site. Moreover, slight deviations in measurement or positioning can compromise implant–abutment fit, alter stress distribution, and potentially lead to biological complications such as peri‑implantitis over the long term~\cite{choi2023influence}. These limitations underscore the need for more automated and reliable approaches to abutment design.

Although computer-aided design (CAD) methods have been proposed to improve the efficiency of abutment design, they still rely heavily on manual operation~\cite{al2022fundamentals}. Recent advances in deep learning have significantly promoted intelligent modeling in digital dentistry~\cite{tian2021efficient,shen2023transdfnet, jang2021fully,cui2021tsegnet,yang2024two,yang2024implantformer}, enabling data‑driven analysis and geometric understanding of complex oral structures. These developments demonstrate the feasibility of learning anatomical representations directly from intraoral scans and provide a foundation for automated treatment planning~\cite{hosseinimanesh2025personalized,qiu2022darch}.
In the field of abutment design, TCEAD~\cite{zheng2026text} introduced the first semi‑automated framework, which uses CLIP‑based embeddings to incorporate clinical textual descriptions—such as implant position and available stock components—for assisting the prediction of abutment parameters. SSA3D~\cite{zheng2025ssa3d} further integrates text‑conditioned prompts with self‑supervised auxiliary tasks, eliminating separate pre‑training and fine‑tuning stages and significantly improving the accuracy of abutment parameter estimation.
While these approaches raise the level of automation compared to traditional manual workflows, they are primarily designed for single‑tooth scenarios and still requires manually specified implant positions. However, in clinical practice, most dental implant cases involve multiple implants. 
The above methods require repeated site‑specific inferences and show limited sensitivity to subtle positional differences between adjacent teeth, which can lead to ambiguous predictions. Moreover, the need for dentists to manually locate implant sites considerably limits workflow efficiency. An ideal automated solution should minimize manual intervention.
Motivated by these considerations, we aim to develop a unified, position‑aware framework that enables fully automated and scalable design for multiple abutments.

In this paper, we propose TEMAD, a text‑conditioned multi‑expert regression framework for automated multi‑abutment design. Given an intraoral scan as input, TEMAD simultaneously localizes all implant sites and predicts the corresponding abutment parameters in a single forward pass, enabling coordinated optimization across multiple implant regions.
Specifically, TEMAD consists of two core components: an Implant Site Identification Network (ISIN) and a Multi‑Abutment Regression Network (MARN). ISIN is a point‑cloud classification network that locates dental implant sites in the patient’s scan and forwards the identified site information to MARN for abutment parameter prediction. Within MARN, we design a Tooth‑Conditioned Feature‑wise Linear Modulation (TC‑FiLM) module and a System‑Prompted Mixture‑of‑Experts (SPMoE) module. TC‑FiLM integrates textual position descriptions with geometry‑aware 3D features extracted from the oral scan, guiding the network to focus on distinct implant regions and enhancing discrimination between anatomically adjacent sites. SPMoE then uses the output features of TC‑FiLM to select a dedicated regression expert based on the specified implant system. This design enables system‑aware parameter prediction and constrains the outputs within clinically valid ranges.
To enhance the network’s ability to capture oral anatomical geometry, we leverage a large‑scale collection of unlabeled intraoral scan data and employ a masked autoencoder to pre‑train the feature encoder. 
We constructed a dataset encompassing both single‑implant and multi‑implant cases to evaluate the effectiveness of TEMAD.
In summary, our contributions are threefold:
\begin{itemize}
\item We propose TEMAD, the first text‑conditioned multi‑abutment regression framework that integrates implant site localization with system‑aware abutment parameter prediction in a fully automated pipeline.
\item We design a Tooth‑Conditioned Feature‑wise Linear Modulation (TC‑FiLM) module, which fuses textual position descriptions with geometry‑aware 3D features from the oral scan. This guides the network to focus on distinct implant regions and enhances discrimination between adjacent anatomical sites.
\item We design a System‑Prompted Mixture‑of‑Experts (SPMoE) module. It selects a dedicated regression expert according to the specified implant system, enabling system‑aware prediction while constraining outputs within clinically valid ranges.
\item Extensive experiments on both single‑ and multi‑abutment datasets show that TEMAD achieves state‑of‑the‑art performance.
\end{itemize}

% In summary, our contributions are threefold:
% \begin{itemize}
	% \item We propose TEMAD, a unified text-conditioned multi-expert regression framework that integrates edentulous site localization and system-aware abutment generation into a fully automated multi-abutment pipeline.
	% \item We introduce a Tooth-Conditioned FiLM module for position-aware feature modulation and a System-Prompted MoE module to enforce implant system–compatible parametric regression.
	% \item Extensive experiments on both single- and multi-abutment datasets demonstrate superior accuracy and improved clinical compatibility.

\section{Related Work}
\subsection{Deep Learning for Dental Structure Analysis}
% Deep learning has significantly advanced automated analysis of dental structures by enabling data-driven extraction of anatomical and pathological features from dental imaging data. For example, Sivari et al.~\cite{sivari2025apd} proposed APD-FFNet, an explainable deep learning framework for automated periodontitis diagnosis from panoramic radiographs. In restorative dentistry, Hosseinimanesh et al.~\cite{hosseinimanesh2025personalized} developed a point-to-mesh completion network to generate personalized dental crown meshes from 3D point clouds of prepared teeth. For tooth segmentation tasks, Xi et al.~\cite{xi20253d} introduced CrossTooth, a boundary-aware deep network for accurate 3D tooth segmentation from intraoral scans. Deep learning has also been explored for implant planning; Yang et al.~\cite{yang2024two} proposed the TSIPR framework to predict implant positions using multi-scale geometric feature learning.  
% These studies demonstrate the strong capability of deep learning in modeling complex oral anatomy and extracting clinically relevant geometric features, providing a foundation for automated implant-related analysis and design.

Deep learning has significantly advanced automated analysis of dental structures by enabling data-driven modeling of complex oral anatomy from 3D scan data. In general, intraoral scans are represented in two major forms, i.e., point clouds and polygonal meshes.
Point-based methods have been widely adopted due to their flexibility in handling irregular and sparse scan data~\cite{xi20253d}. For example, Cui et al.~\cite{cui2021tsegnet} proposed TSegNet for tooth segmentation on dental point clouds, introducing distance-aware centroid prediction and a confidence-guided cascade segmentation module to achieve efficient and accurate tooth delineation. Meanwhile, mesh-based learning approaches explicitly model surface connectivity and geometric topology~\cite{wu2022two}. Wang et al.~\cite{wang2024weakly} developed the WS-TIS framework to learn discriminative mesh features for automatic tooth surface labeling from raw intraoral scans, demonstrating the effectiveness of mesh representations in capturing structural relationships.
However, supervised learning for 3D dental analysis typically requires dense point- or face-level annotations, which are expensive and time-consuming to obtain. To alleviate this limitation, self-supervised learning has recently been introduced for oral scan understanding. Almalki et al.~\cite{almalki2024self} proposed DentalMAE, which integrates masked autoencoding with self-supervised pretraining on large-scale unlabeled dental meshes. This strategy significantly improves generalization ability and segmentation accuracy in downstream tasks~\cite{krenmayr2025evaluating}, highlighting the potential of representation pretraining for dental applications.

Beyond structural understanding tasks such as segmentation, deep learning has also been explored in broader clinical workflows~\cite{sivari2025apd, yang2024two,yang2024implantformer}. 
Zhuang et al. proposed a hybrid framework integrating semantic–instance segmentation and pose alignment to enable robust automatic tooth segmentation and fine labeling~\cite{zhuang2023robust}. 
In restorative dentistry, Hosseinimanesh et al.~\cite{hosseinimanesh2025personalized} developed a point-to-mesh completion network to generate personalized crown geometries from prepared tooth scans. 
Wu et al. developed a three-stage pipeline that performs defect completion, image enhancement, and point-to-mesh reconstruction to generate personalized 3D dental models from raw observations~\cite{wu2025automatic}. 
These studies collectively demonstrate the growing capability of deep learning in supporting automated implant-related analysis and design, motivating the development of unified frameworks that can directly infer clinically meaningful geometric parameters from complex oral scan data.

\subsection{Traditional Abutment Design}
Measuring the restoration space is a crucial step in implant abutment design, as it directly affects the fit and stability of the implanted tooth within the patient's oral structure~\cite{abichandani2013abutment}.
In traditional clinical workflows, dentists first acquire intraoral scans or physical impressions and construct gingival models, followed by manual measurement of several parameters, including the transgingival thickness, implant diameter, and gingival–occlusal height~\cite{shah2023literature}.
Based on these measurements and prior implantation systems, clinicians need to select a compatible stock abutment for installation and verification.
This process is highly labor-intensive and often requires multiple iterations of fitting and adjustment.
To improve efficiency, CAD technologies were introduced to assist restoration-space measurement and abutment design on digital models~\cite{priest2005virtual}. Subsequent CAD–CAM systems enable the fabrication of customized abutments with higher precision and personalization~\cite{zhang2017method,kang2020abutment}. However, these approaches still rely heavily on manual identification of measurement positions and expert experience. Different measurement locations on the gingival contour can lead to inconsistent parameter estimates, potentially resulting in inappropriate abutment selection and long-term biological complications such as peri-implantitis~\cite{prpic2025influence}. Moreover, in multi-abutment cases, these traditional workflows typically repeat the single-abutment design process for each implant site, making the overall procedure complex and inefficient when multiple restorations are required within the same oral cavity.

\subsection{Deep Learning-Based Automated Abutment Design}
Recent advances in deep learning have begun to enable automated prediction of abutment design parameters directly from intraoral scans. Zheng et al.~\cite{zheng2026text} proposed TCEAD, which integrates MeshMAE pre-training with CLIP-based text guidance to localize implant sites and regress abutment parameters. This framework demonstrates strong performance in single-tooth abutment design tasks and represents an early attempt toward data-driven abutment design. To alleviate the limitation of limited annotated data, Zheng et al.~\cite{zheng2025ssa3d} further introduced SS$A^3$D, which incorporates a self-supervised auxiliary reconstruction task and a text-conditioned prompt module to improve representation learning while reducing the need for large-scale labeled data.
Despite these advances, existing deep learning-based approaches remain limited to single-tooth scenarios and do not provide a fully automated workflow. They typically require manual specification of edentulous tooth positions for each inference and perform separate predictions for individual implant sites. Consequently, these methods struggle to handle multi-abutment cases and exhibit limited capability in modeling the complex geometric relationships within the oral cavity when multiple missing teeth are present.

\section{Method}
\begin{figure}\centering
\includegraphics[scale=0.3]{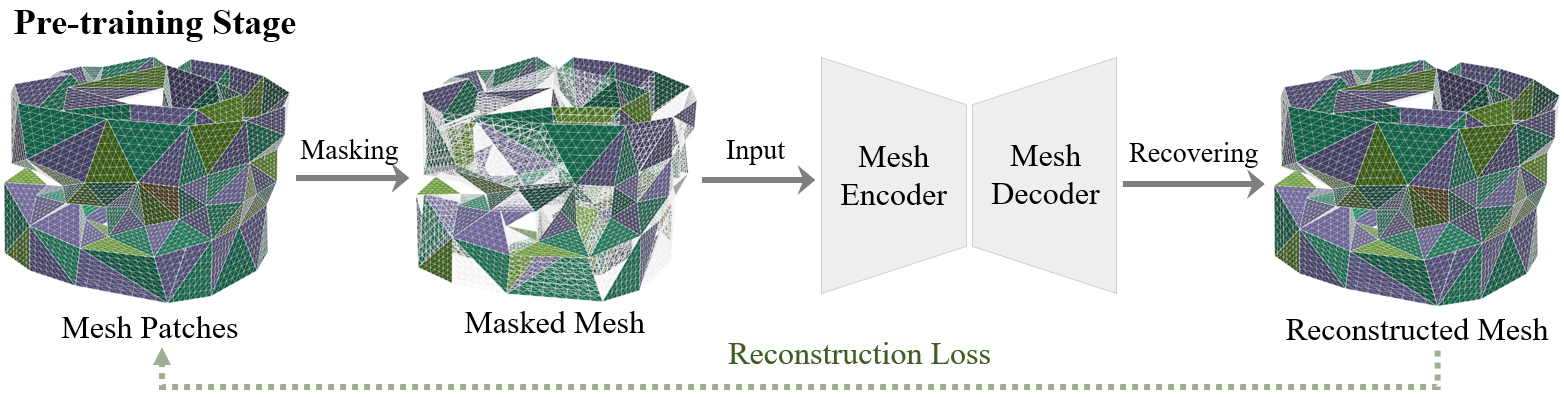}
\caption{Illustration of the self-supervised masked autoencoder pretraining framework.}
\label{pretraining}
\end{figure}

\begin{figure*}\centering
\includegraphics[scale=0.5]{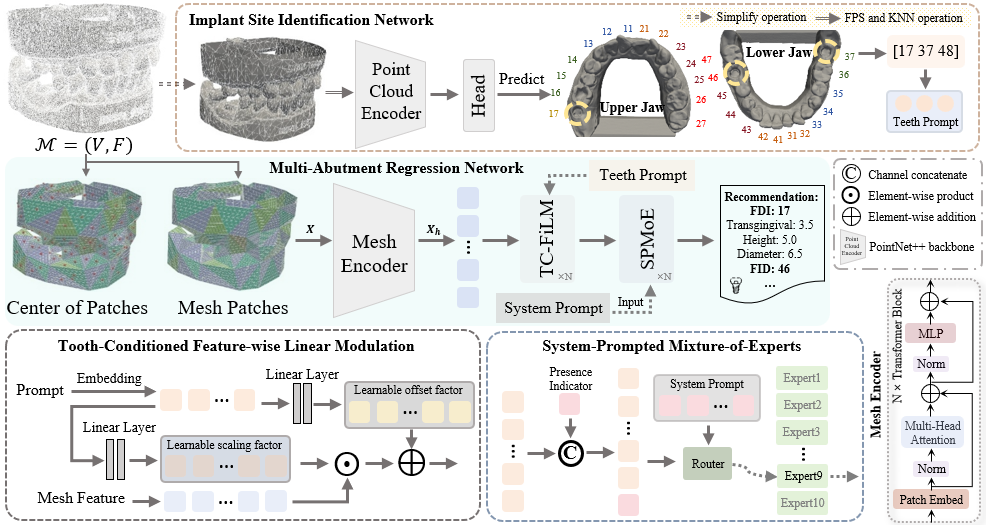}
\caption{The overall framework of the proposed method.}
\label{Overall_Framework}
\end{figure*}

% Given a patient’s intraoral scan and a clinician-specified implant system, TEMAD aims to automatically generate system-compliant abutment parameters for all edentulous sites in a single end-to-end forward pass. An overview of the proposed framework is illustrated in Fig.~\ref{Overall_Framework}.
% TEMAD is composed of two key networks: an Implant Site Identification Network (ISIN) for perceiving edentulous tooth positions and a Multi-Abutment Regression Network (MARN) for predicting corresponding abutment parameters. To improve training stability, the two networks are trained separately. During inference, the intraoral scan mesh is fed to TEMAD, within which the ISIN first predicts the locations of edentulous tooth positions. The predicted site identities are subsequently incorporated as auxiliary prompt into the MARN to facilitate slot-aware parameter regression for multiple implant sites. 
% Detailed descriptions of these components are provided in the following sections.
Given a patient’s intraoral scan and a clinician‑specified implant system, TEMAD automatically predicts the abutment parameters for all implant sites in a single forward process. An overview of the proposed framework is illustrated in Fig.~\ref{Overall_Framework}.
TEMAD comprises two key networks: an Implant Site Identification Network (ISIN) that detects implant sites, and a Multi‑Abutment Regression Network (MARN) that predicts the corresponding abutment parameters. We first train ISIN independently and then integrate it as a fixed sub‑network into MARN.
During the training of MARN, ISIN first predicts the locations of implant sites. These predicted sites are then incorporated as auxiliary prompts into MARN to complete the prediction of multiple abutment parameters. 
In addition, to further enhance the geometric representation capability of the encoder, we pretrain it with a masked autoencoder on a large‑scale collection of unlabeled intraoral scans. The overall workflow, including the pre‑training stage, is depicted in Fig.~\ref{pretraining}.
Detailed descriptions of each component are provided in the following sections.

\subsection{Self-Supervised Encoder Pretraining} \label{ssl}
% To improve the encoder’s ability to model complex 3D oral anatomy, we perform self-supervised pretraining on intraoral scan meshes. Due to the irregular geometry and limited annotations in dental data, directly learning task-specific features is challenging. By adopting a masked reconstruction strategy, the encoder is encouraged to capture intrinsic spatial dependencies and global structural patterns.
To overcome the large structural differences of intraoral scans and the limited availability of annotations, we conduct self-supervised pretraining on large-scale oral mesh datasets. Masked autoencoding enables the encoder to learn transferable representations by reconstructing heavily masked regions from visible context, thereby capturing global anatomical priors and long-range spatial dependencies. The pretrained encoder provides geometry-aware initialization for downstream tasks, which significantly eases optimization in the complex latent design space of multi-abutment parameter prediction and improves overall prediction robustness.

\textbf{Patch-Based Mesh Encoding.}
Given an oral scan mesh $\mathcal{M}=(V, F)$, we first normalize its resolution and partition it into regular patches following~\cite{hu2022subdivision,zheng2026text}. 
Each patch contains a fixed number of faces. Let $F_n$ denote the number of faces within a patch. For each face, we extract a 13-dimensional geometric descriptor and represent the patch as a set of face-level features $E=\{e_i\}_{i=1}^{F_n}$, which are embedded into latent tokens via an MLP. To preserve spatial structure, positional embeddings $P=\{p_i\}_{i=1}^{F_n}$ derived from patch centroids are further added to the embedded patch tokens, yielding the final token representation.
The resulting patch tokens $X=\{x_i\}_{i=1}^{F_n}$ are fed into a transformer-based encoder to capture long-range geometric dependencies across the oral mesh. 

\subsubsection{Encoder–Decoder Architecture.}
During self-supervised pretraining, we adopt an encoder–decoder architecture to learn robust geometric representations from oral meshes. The encoder follows the same transformer backbone used in the downstream network and consists of 12 standard transformer blocks that process the visible patch tokens with positional embeddings. After encoding, a lightweight transformer decoder composed of 6 transformer blocks is employed to reconstruct the masked patches. The decoder takes both the encoded visible tokens and shared mask embeddings as input, while positional information is added to all tokens to preserve spatial consistency across patches. The reconstructed features of the masked regions are then used to guide representation learning through a reconstruction objective.

\subsubsection{Reconstruction Objective.}
% The reconstruction task supervises both vertex geometry and face features. For each patch, the Chamfer distance is computed between reconstructed vertex relative coordinates $V_p$ and ground-truth coordinates $G_q$:
The reconstruction task supervises both vertex geometry and face features at the patch level. For each masked patch, the decoder predicts a set of relative vertex coordinates (relative to the center point of the patch) $V_p=\{p_i\}_{i=1}^{45}$, where 45 denotes the fixed number of vertices within a patch. 
The ground-truth vertex set is similarly defined as $G_q=\{q_i\}_{i=1}^{45}$, representing the relative coordinates of the corresponding patch vertices with respect to the patch centroid.
To measure geometric reconstruction quality, the Chamfer distance is computed between the predicted vertex set $V_p$ and the ground-truth set $G_q$:
\begin{equation}
	L_{geo}(V_p, G_q) = \frac{1}{|V_p|}\sum_{p \in V_p}\min_{q \in G_q} ||p - q||_2 +\frac{1}{|G_q|}\sum_{q \in G_q} \min_{p \in V_p} ||q-p||_2.
\end{equation}
An additional MLP head predicts face features $\hat{h}_i$, supervised by mean squared error,
\begin{equation}
	L_{feat}= \frac{1}{|F|} \sum_{i\in F} \| h_i - \hat{h}_i \|_2^2
\end{equation}
where $h_i$ and $\hat{h}_i$ denote ground-truth and predicted face features, respectively. The overall reconstruction loss is defined as:
\begin{equation}
	L_{rec} = L_{geo} + \zeta L_{feat},
\end{equation}
where $\zeta$ is the loss weight of face features.

\subsection{Implant Site Identification Network}
Given a high-resolution intraoral mesh $\mathcal{M}$, we simplify it into a lightweight point representation with 9000 faces for efficient inference. Specifically, we uniformly sample $V=4096$ vertices to obtain a point set $\mathcal{P}=\{x_i \in \mathbb{R}^3\}_{i=1}^N$, preserving global dental arch geometry while reducing computational cost.
Implant site localization can be defined as a multi-label classification problem over anatomically defined tooth positions. Following the Fédération Dentaire Internationale (FDI) tooth numbering system, $\mathcal{B}$ $(\mathcal{B}=28)$ permanent tooth sites (excluding third molars) are considered. The module predicts a binary vector $z \in \{0,1\}^{\mathcal{B}}$, where $z_k=1$ indicates that site $k$ is implant site.

To model both local socket geometry and global dental arch context from the downsampled point cloud, we evaluate three representative point-based architectures covering point-wise feature learning, hierarchical neighborhood aggregation, and transformer-based modeling. Based on extensive empirical evaluation, we observe that PointNet++ consistently achieves superior localization accuracy while maintaining a lightweight architecture, with significantly lower parameter counts and FLOPs compared to alternative designs. Therefore, as validated by the ablation study in Sec.~\ref{experimental_analysis}, we adopt a PointNet++ backbone with multi-scale grouping (MSG) to extract hierarchical geometric features,
% A PointNet++ backbone with multi-scale grouping (MSG) is employed to encode geometric features, 
\begin{equation}
f = \Phi(\mathcal{P}) \in \mathbb{R}^{D},
\end{equation}
where $\Phi(\cdot)$ denotes three hierarchical set abstraction layers. The global feature $f$ is fed into a three-layer MLP with sigmoid activation to produce predicted probabilities $\hat{\mathbf{z}} \in [0,1]^{\mathcal{B}}$.
The module is trained using binary cross-entropy loss,
\begin{equation}
\mathcal{L}_{\text{ISIN}} =
-\frac{1}{\mathcal{B}} \sum_{k=1}^{\mathcal{B}}
\left[
z_k \log \hat{z}_k
+
(1 - z_k)\log(1 - \hat{z}_k)
\right].
\end{equation}

During inference, sites with $\hat{z}_k > \tau$ ($\tau=0.5$) are identified as implant site. The predicted implant site set is subsequently provided to the Tooth-Conditioned FiLM module to constrain downstream abutment regression.

\subsection{Multi-Abutment Regression Network}
To enable simultaneous parameter prediction for multiple implant sites within a single oral scan, we design a unified multi-abutment regression network. The network is composed of three key components: 1) Mesh Encoder, 2) Tooth-Conditioned Feature-wise Linear Modulation (TC-FiLM), and 3) System-Prompted Mixture-of-Experts (SPMoE). 
To obtain robust geometric representations while preserving mesh topology, we adopt a masked autoencoding framework based on MeshMAE~\cite{liang2022meshmae}.
Next, we will introduce these modules in detail. 

% To enable simultaneous parameter prediction for multiple implant sites within a single oral scan, we design a unified multi-abutment regression network. The network is composed of three key components: (1) Mesh Encoder, (2) Tooth-Conditioned Feature-wise Linear Modulation (TC-FiLM), and (3) System-Prompted Mixture-of-Experts (SPMoE). These modules are introduced in detail in the following subsections. The overall network is trained using a robust regression objective for reliable multi-site parameter estimation.

% \subsubsection{Geometry-Aware Representation Learning}
% To obtain robust geometric representations while preserving mesh topology, we adopt a masked autoencoding framework based on MeshMAE~\cite{liang2022meshmae}. The encoder is pre-trained on large-scale oral scan meshes and subsequently transferred to the downstream regression task.

% \textbf{Patch-Based Mesh Encoding.}
% Given an oral scan mesh $\mathcal{M}=(V, F)$, we first normalize its resolution and partition it into regular patches following~\cite{hu2022subdivision,zheng2025text}. 
% Each patch $E=\{e_i\}_{i=1}^{F_n}$ is represented by 13-dimensional face-level geometric features and embedded via an MLP. Positional embeddings $P=\{p_i\}_{i=1}^{F_n}$ derived from patch centroids are added to obtain the final token representation.  

\textbf{Mesh Encoder.}
% The resulting patch tokens are fed into a transformer-based encoder to capture long-range geometric dependencies across the oral mesh. 
The encoder consists of 12 standard transformer blocks, each comprising multi-head self-attention and feed-forward layers, as illustrated in Fig.~\ref{Overall_Framework}. Through stacked self-attention operations, the network progressively aggregates contextual information from neighboring and distant patches, enabling the modeling of global anatomical structures and spatial relationships among teeth. 
To enhance the robustness of the representation, the encoder is initialized with weights obtained from a self-supervised masked pretraining scheme (Section~\ref{ssl}), which facilitates the extraction of informative structural features of the oral cavity.

%The encoder comprises 12 standard Transformer blocks. 
%During pre-training, a random subset of mesh patches is masked, and a Transformer encoder processes the visible tokens. A lightweight decoder reconstructs the masked geometry and features. 
%After pre-training, the decoder is discarded and the encoder is used to extract geometry-aware representations for abutment parameter regression.

\begin{figure*}\centering
\includegraphics[scale=0.32]{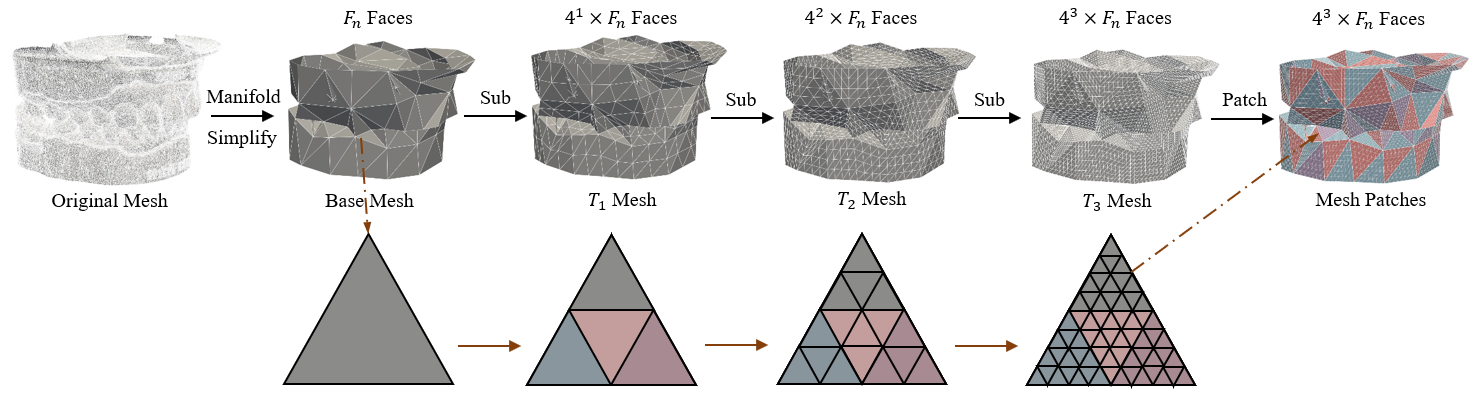}
\caption{Visualization of the remeshing process. The original intraoral scan mesh is first converted into a base mesh through manifold repair and mesh simplification. The base mesh is then refined through three successive subdivision steps, where each face is split into four ordered faces at each step. Finally, the refined mesh is partitioned into $F_n$ patches, where each patch contains 64 faces generated after three subdivisions, serving as the input to the network.}
\label{remesh_process}
\end{figure*}

\subsubsection{Tooth-Conditioned Feature-wise Linear Modulation}
In multi-abutment scenarios, each edentulous site should be processed independently to prevent interference between anatomically distinct tooth regions. To achieve position-specific modulation, we introduce Tooth-Conditioned Feature-wise Linear Modulation (TC-FiLM), which generates slot-specific affine transformations conditioned on tooth identity.
For each predicted implant site $k \in S_{\text{implant}}$, a learnable embedding is defined as
\begin{equation}
\mathbf{e}_k = \mathbf{W}_{\text{emb}}[k] \in \mathbb{R}^{C},
\end{equation}
where $\mathbf{W}_{\text{emb}} \in \mathbb{R}^{\mathcal{B} \times C}$ is a trainable embedding table encoding anatomical priors for all tooth positions.
The embedding $\mathbf{e}_k$ is projected to modulation parameters via linear layers:
\begin{equation}
\boldsymbol{\gamma}_k =  W_{\gamma} \mathbf{e}_k + \mathbf{b}_{\gamma}, 
\quad
\boldsymbol{\beta}_k = W_{\beta} \mathbf{e}_k + \mathbf{b}_{\beta},
\end{equation}
where $W_{\gamma}, W_{\beta} \in \mathbb{R}^{C \times H}$ are learnable matrices.

Given the mesh feature $X_h \in \mathbb{R}^{H}$ from the encoder, the modulated feature for site $k$ is computed as
\begin{equation}
X_k = X_h \odot (1 + \boldsymbol{\gamma}_k) + \boldsymbol{\beta}_k,
\end{equation}
where $\odot$ denotes element-wise multiplication.
This method enables tooth-specific feature adaptation by applying independent affine transformations for each implant site, thereby supporting scalable multi-abutment regression without cross-site interference.

\subsubsection{System-Prompted Mixture-of-Experts}
In clinical practice, geometric parameters of the abutment is constrained by the selected implant system, as different manufacturers impose distinct platform dimensions and emergence profiles. To model system-specific parameter spaces, we introduce a System-Prompted Mixture-of-Experts (SPMoE) module.
\par
For each case, we consider up to $T=3$ candidate implant sites. A binary presence indicator $m_i \in \{0,1\}$ denotes whether slot $i$ corresponds to a predicted implant site. Given the tooth-conditioned feature $h_i \in \mathbb{R}^{H}$ from TC-FiLM, we concatenate the indicator to obtain
\begin{equation}
u_i = [h_i ; m_i] \in \mathbb{R}^{H+1}.
\end{equation}
This explicit validity encoding allows the regressor to distinguish active and inactive slots without additional masking.
\par
Let $c \in \{0, \dots, C-1\}$ denote the implant system identifier ($C=13$ in our implementation). We instantiate $C$ independent regression experts $\{\mathcal{E}_c(\cdot)\}_{c=0}^{C-1}$, each implemented as a two-layer MLP.
Expert selection is deterministically guided by the system prompt:
\begin{equation}
\mathbf{o}_i =
\begin{cases}
\mathcal{E}_c(\mathbf{u}_i), & \text{if } m_i = 1, \\
\mathbf{0}, & \text{if } m_i = 0.
\end{cases}
\end{equation}
\par
The proposed SPMoE module maintains the independence of implant sites among multiple implant sites while achieving system-aware regression.

\subsubsection{Regression Objective}
Abutment parameter regression is supervised using a weighted combination of mean squared error (MSE) and Smooth L1 loss to balance stability and robustness to outliers.
For each active slot $i$ with presence indicator $m_i = 1$, let $y_i$ and $y_i^{gt}$ denote the predicted and ground-truth abutment parameter vectors, respectively. The regression loss is defined as
\begin{equation}
\mathcal{L}_{\text{reg}} =
\frac{1}{\sum_i m_i}
\sum_{i} m_i 
\left(
\alpha \, \mathcal{L}_{\text{MSE}}(y_i, y_i^{gt})
+
\beta \, \mathcal{L}_{\text{SL1}}(y_i, y_i^{gt})
\right),
\end{equation}
where $\alpha$ and $\beta$ are balancing coefficients.
The MSE term is defined as
\begin{equation}
\mathcal{L}_{\text{MSE}}(y_i, y_i^{gt})
=
\frac{1}{d}
\left\|
y_i - y_i^{gt}
\right\|_2^2,
\end{equation}
where $d$ denotes the dimensionality of the parameter vector.
The Smooth L1 loss is defined element-wise as
\begin{equation}
\mathcal{L}_{\text{SL1}}(u) =
\begin{cases}
0.5\cdot u^2/\delta, & |u| < \delta, \\
|u| - 0.5 \delta, & \text{otherwise},
\end{cases}
\end{equation}
where $u = y_i - y_i^{gt}$ and $\delta$ controls the transition point between quadratic and linear regions.

\begin{figure*}[htbp]
    \centering
    \subfigure{}{\includegraphics[width=0.32\textwidth]{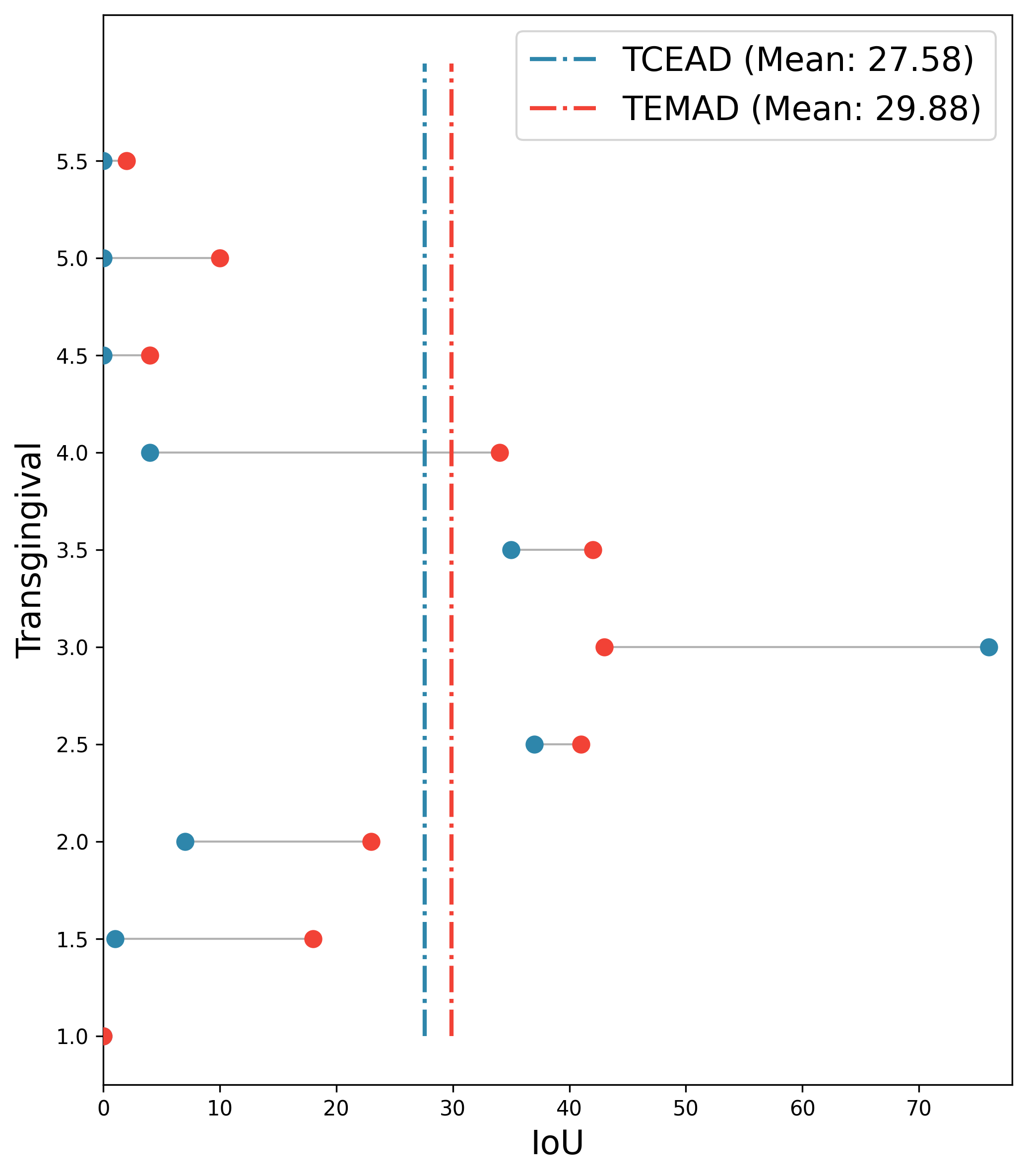}}
    \subfigure{}{\includegraphics[width=0.32\textwidth]{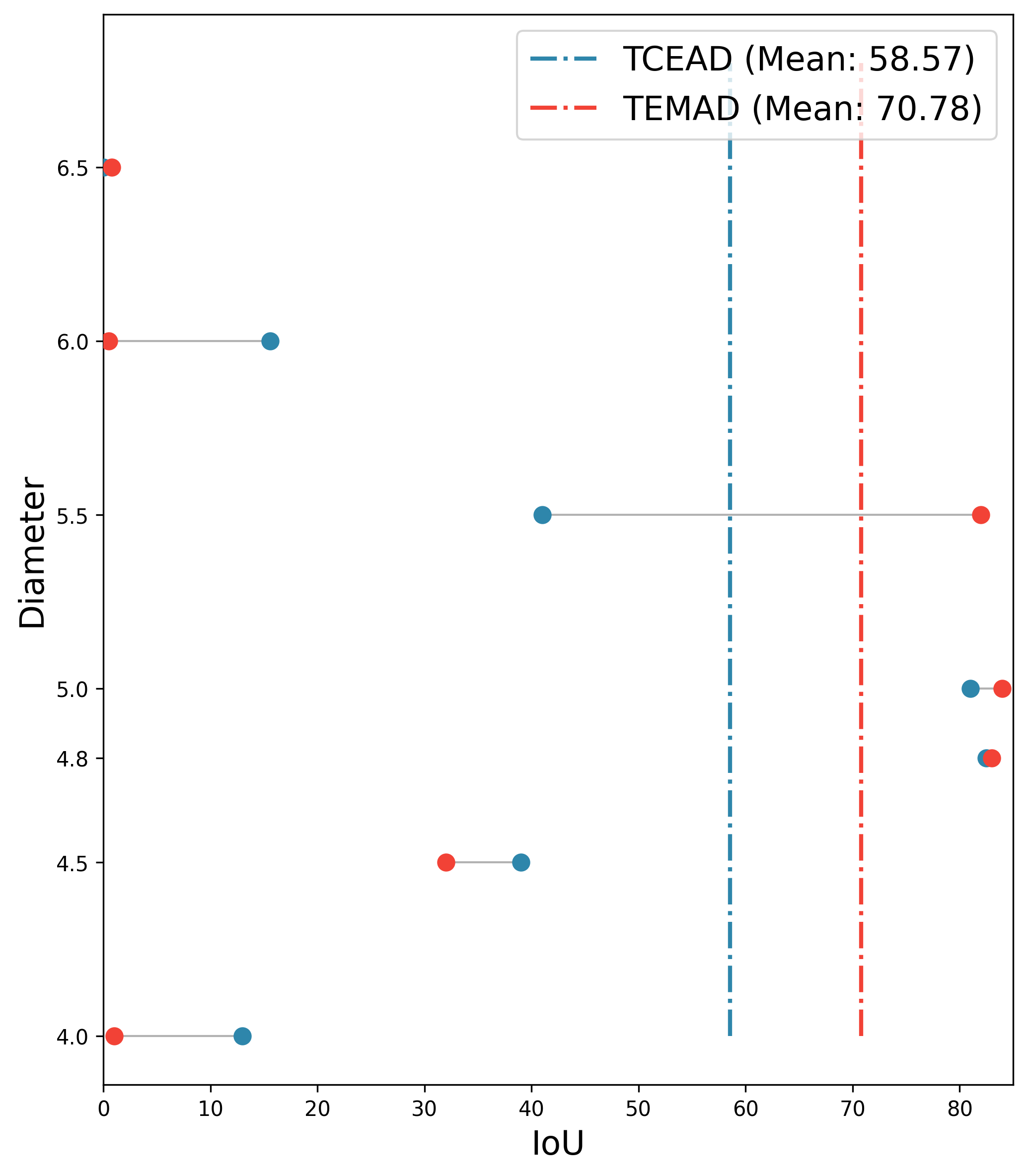}}
    \subfigure{}{\includegraphics[width=0.32\textwidth]{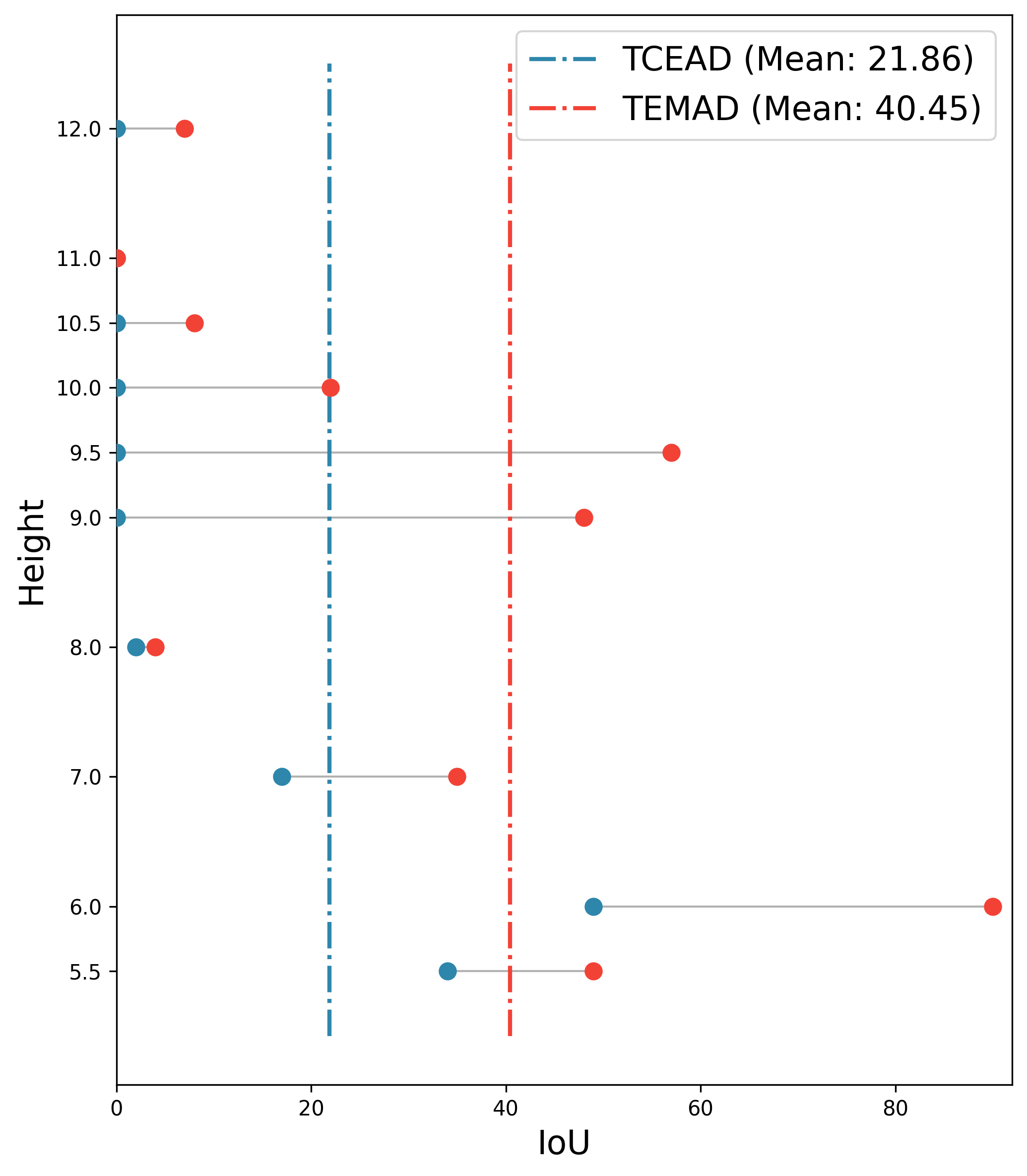}}
    \caption{Comparison of parameter-specific performance for multi-abutment prediction. Each subplot corresponds to one key parameter. The horizontal axis denotes IoU performance, while the vertical axis lists the values of the corresponding parameter.}
    \label{Multi_Compare}
\end{figure*}

\begin{figure*}\centering
\includegraphics[scale=0.30]{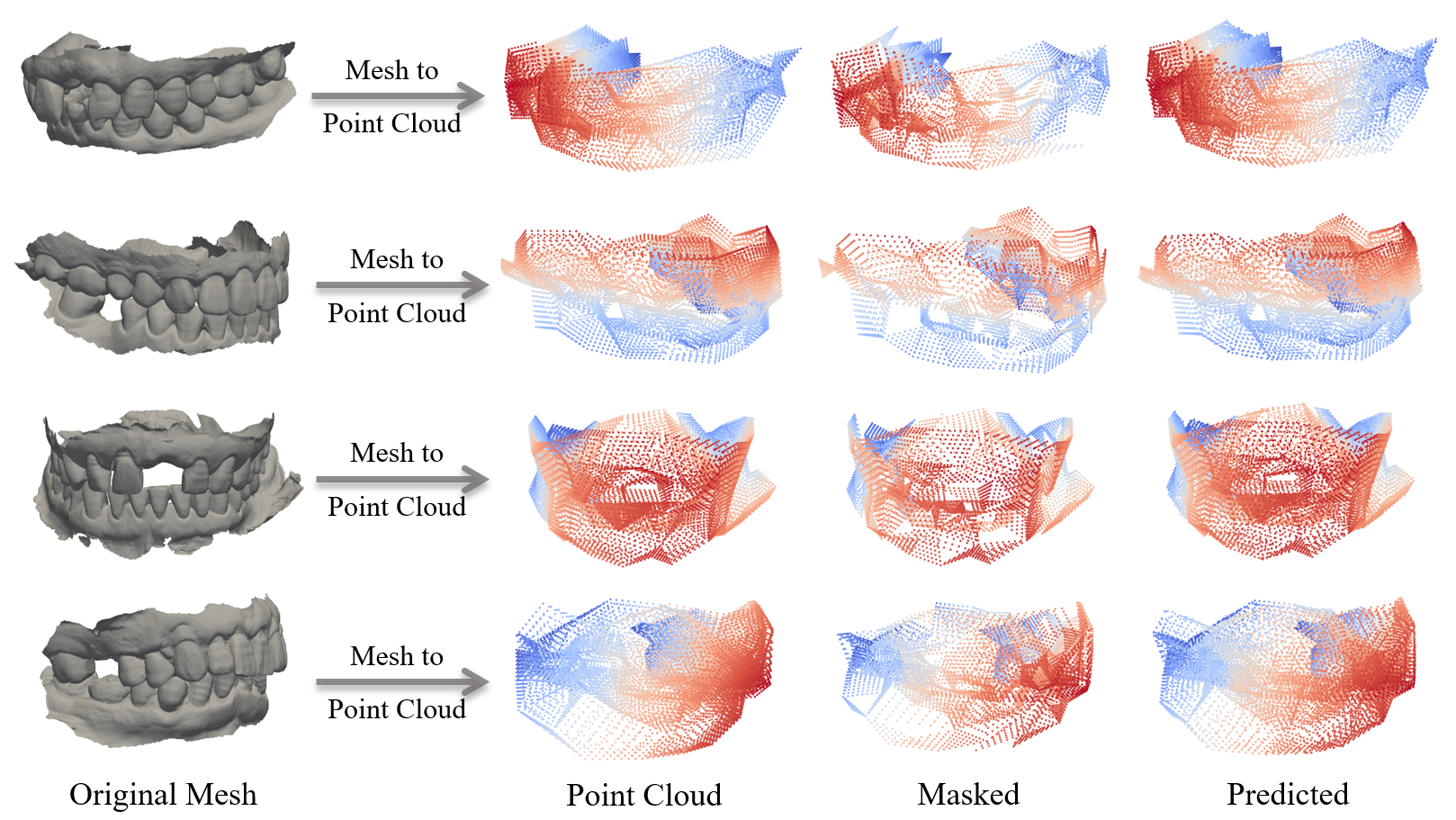}
\caption{Visualization of pre-trained reconstruction performance.}
\label{reconstruction_vis}
\end{figure*}

\section{Experiments and Results}

\subsection{Dataset}

\textbf{Dataset.}
We construct a large-scale intraoral mesh dataset with 22,115 scans for self-supervised pre-training, including 933 scans from Teeth3Ds dataset~\cite{ben2022teeth3ds+}, 2,121 scans from Orthodontic dataset, and 19,061 additional unlabeled scans. 
For supervised learning, we curate a proprietary annotated dataset with three clinically measured abutment parameters per case. The training set contains 9,037 single-abutment and 1,688 multi-abutment scans, while the test set includes 1,211 single-abutment and 296 multi-abutment scans. The same split is used for ISIN training and evaluation.

\textbf{Preprocessing.}
All meshes undergo standard manifold cleaning to remove holes and self-intersections. 
As illustrated in Fig.~\ref{remesh_process}, each mesh is first simplified to a base mesh with 500 faces to normalize geometric resolution. The base mesh is then progressively refined through three successive MAPS subdivision~\cite{hu2022subdivision} steps, resulting in a structured mesh with approximately 32,000 faces. This hierarchical remeshing process enables regular patch partitioning and facilitates stable geometry-aware feature extraction in the subsequent network.

\subsection{Implementation Details}
% For self-supervised pre-training, we use AdamW with a batch size of 64 and an initial learning rate of $1\times10^{-4}$ under cosine decay for 500 epochs. 
% Fine-tuning uses a batch size of 32 with the same optimizer and learning rate for 100 epochs. Data augmentation includes random scaling, rotation, and shape deformation. 
All models are implemented in PyTorch and trained on the NVIDIA A40 GPU platform. For self-supervised pre-training, the mesh encoder is optimized using the AdamW optimizer with a batch size of 64 and an initial learning rate of $1\times10^{-4}$, following a cosine decay schedule. The network is trained for 500 epochs with a masking ratio of 50\%. A weighted reconstruction loss is employed, where the face feature reconstruction term $\zeta$ is assigned a weight of 0.5. Data augmentation strategies include random scaling, random rotation, and shape deformation to improve representation robustness.
During fine-tuning, the network is trained for 100 epochs using the same optimizer and initial learning rate, with a reduced batch size of 32. The same data augmentation methods are applied as in the pre-training stage. For abutment parameter regression, the loss function is defined as a weighted combination of mean squared error and Smooth L1 loss, where the balancing coefficients are set to $\alpha = 1$ and $\beta = 5$.

\subsection{Evaluation Criteria}
In clinical practice, abutment selection is determined by matching key geometric parameters to stock components within a tolerance range. We therefore evaluate regression accuracy using an Intersection over Union (IoU) metric under a clinically acceptable tolerance interval.

Each predicted parameter is represented as an interval $[y_i - \mu,\, y_i + \mu]$, and IoU with the ground truth is computed as:

\begin{equation}
IoU(y_i, y_i^{gt}) =
\frac{|[y_i-\mu, y_i+\mu] \cap [y_i^{gt}-\mu, y_i^{gt}+\mu]|}
{|[y_i-\mu, y_i+\mu] \cup [y_i^{gt}-\mu, y_i^{gt}+\mu]|}
\end{equation}
where $y_i$ and $y_i^{gt}$ denote predicted and ground-truth values, and $\mu = 1.0$ mm defines the tolerance. 
Under this formulation, an IoU of 0.33 corresponds to an absolute error of approximately 0.5 mm, which satisfies typical clinical requirements. 
Evaluation is performed only on correctly identified implant sites.

\subsection{Performance Analysis}\label{experimental_analysis}
\paragraph{\textbf{Ablation Study of ISIN}}
ISIN is used to provide implant site locations for the abutment parameter regression network, thus requiring a good trade-off between accuracy and efficiency. Since the number of teeth in the oral cavity is finite, we defined implant site prediction as a classification task based on international FDI tooth alignment. We compared different point cloud classification networks, and the results are shown in Table~\ref{isin_module}.

Among the evaluated models, PointNet++ achieves significantly higher accuracy in both implant site localization ($A_{ISL}$) and implant site count prediction ($A_{ISC}$), while maintaining relatively low parameter size and computational cost. 
Here, $A_{ISL}$ is defined as a strict case-level metric, where a prediction is considered correct only if all edentulous sites within a given oral scan are correctly identified; partial matches or over-predictions are regarded as incorrect. 
To provide a more tolerant evaluation of localization performance, we further report distance-based metrics (Acc@k), which measure the proportion of ground-truth implant sites whose predicted positions fall within k neighboring tooth locations. 
%These metrics relax the strict matching requirement of $A_{ISL}$ and better reflect the model's ability to approximate implant positions.
From the results, PointNet++ achieves the best performance across all Acc@k metrics (92.76\%, 92.90\%, and 92.93\% for Acc@1, Acc@2, and Acc@3), indicating that most predictions are either exact or within one adjacent tooth, demonstrating strong spatial localization capability. 
PointNet shows moderate improvement under relaxed criteria, suggesting that many predictions are close but not exact. 
In contrast, PointFormer performs worst across all metrics, reflecting both poor exact localization and weak spatial approximation.
Notably, the small gap between Acc@1, Acc@2, and Acc@3 for all models suggests that most localization errors are either very small (within one tooth) or significantly incorrect, with limited intermediate deviations. 
Combined with its superior $A_{ISL}$ and $A_{ISC}$, PointNet++ achieves the best balance between precise and tolerant localization, making it a suitable backbone for ISIN.
%Although PointNet exhibits the smallest model size and comparable FLOPs, its localization accuracy is substantially lower, with an $A_{ISL}$ gap of 49.26\% compared to PointNet++. In contrast, PointFormer shows the lowest prediction accuracy among the candidates while also incurring the highest computational overhead. This superior accuracy–efficiency trade-off makes PointNet++ a suitable backbone for ISIN in the proposed framework.

We further assess the impact of ISIN on the overall framework. In the ablation setting without ISIN, ground-truth tooth positions are provided as conditioning inputs. When replaced with automatically predicted positions during inference, TEMAD maintains stable and competitive IoU performance (Table~\ref{module_compare}). This indicates that ISIN provides reliable localization while enabling fully automated multi-abutment design without noticeable performance degradation.

\begin{table*}[htbp]  
\centering
\setlength{\tabcolsep}{4pt} % 自动适配列间距
\resizebox{\linewidth}{!}{  % 关键：自适应整行宽度
\begin{tabular}{c|c|c|c|c|c|c|c}
\hline
	Network & Params(M) & FLOPs(G) & $A_{ISL}$(\%) & Acc@1(\%) & Acc@2(\%) & Acc@3(\%) & $A_{ISC}$(\%)  \\
	   \hline 
	   PointFormer & 12.91 & 43.01 & 19.86 & 32.98 & 33.64 & 33.64 & 26.57  \\
	   PointNet & \textbf{1.61} & 8.5 & 33.20 & 51.27 & 52.60 & 52.67 & 48.89  \\
	  PointNet++  & 1.74 & \textbf{8.5} & \textbf{82.46} & \textbf{92.76}& \textbf{92.90} & \textbf{92.93} & \textbf{86.43} \\ \hline
	\end{tabular}
	}
\caption{The performance of IAPM. $A_{ISL}$: Implant Site Localization Accuracy, $A_{ISC}$: Implant Site Count Accuracy, Acc@k denotes the proportion of ground-truth implant sites whose predicted locations fall within k adjacent tooth positions (intra-arch distance), Params: Parameters, FLOPs: Floating Point Operations Per Second.}
	\label{isin_module}
\end{table*}

\paragraph{\textbf{Ablation Study of SPMoE}}
We further conduct ablation experiments by removing the SPMoE module to evaluate its contribution. The quantitative results are reported in Table~\ref{module_compare}. 
% Incorporating the SPMoE module consistently improves IoU performance across parameters, with the most notable gain observed in the height parameter. This improvement indicates that system-prompted expert selection effectively constrains regression outputs within implant system–specific parameter ranges. The results demonstrate the benefit of modeling system-aware regression through expert specialization.
When SPMoE is removed while retaining ISIN, the IoU performance decreases across all parameters. Specifically, the IoU for transgingival shows only a marginal change, whereas diameter and height exhibit substantial drops of 15.15 and 20.54, respectively. This notable degradation suggests that a single shared regression head struggles to model the heterogeneous parameter distributions associated with different implant systems.
By introducing system-prompted expert selection, SPMoE effectively constrains regression outputs within implant system–specific parameter ranges. The observed performance gains demonstrate that expert specialization improves both regression accuracy and prediction stability in multi-abutment scenarios.

\begin{table}[]  
\centering
\begin{tabular}{cc|ccc}
\hline
	ISIN & SPMoE & {Transgingival} & {Diameter} & {Height} \\
	   \hline 
	  \cxmark & \ding{52}  & 29.42 & \textbf{72.00} & 40.01 \\
	 \ding{52} & \cxmark  & 26.24 & 55.63 & 19.91 \\
	\ding{52} & \ding{52} & \textbf{29.88} & 70.78 & \textbf{40.45} \\ \hline
	\end{tabular}
	\label{module_compare}
\caption{Ablation experiments on the ISIN and SPMoE module.}
\end{table}

% \paragraph{\textbf{Ablation Study of Mixture-of-Experts Designed in SPMoE.}}  

\begin{table*}[]
\centering
\begin{tabular}{c|c|c|ccc}
\hline
	Dataset  & Modality & Method & Transgingival & Diameter & Height \\ \hline 
	\multirow{8}{*}{SA} 
	&&  PointNet & 24.62 & 62.17 & 22.83 \\
     && PointNet++  & 26.30 & 57.38 & 15.39 \\
     &Point Cloud&  PointMAE  & 25.10 & 55.83 & 43.16 \\
     &&  PointMamba  & 30.14 & 62.17 & \textbf{46.49} \\
     &&  PointFEMAE & 26.60 & 57.70 & 19.16 \\  \cline{2-6}
     &\multirow{4}{*}{Mesh}
     &  MeshMAE & 28.61 & 62.53 & 18.44 \\
	&&  TCEAD & 28.17 & 61.77 & 23.70 \\
	&&  SS$A^3$D & 30.65 & 70.03 & 34.22 \\
	&&  TEMAD(Ours) & \textbf{32.18} & \textbf{70.77} & 40.01\\ 
%	\cline{1-1}\cline{3-6}
	\hline
     \multirow{3}{*}{SMA} &\multirow{3}{*}{Mesh}
	&  TCEAD & 27.58 & 58.57 & 21.86 \\
	&&  SS$A^3$D & 28.91 & 70.25 & 33.52 \\
	&&  TEMAD(Ours) & \textbf{29.88} & \textbf{70.78} & \textbf{40.45} \\ \hline
	\end{tabular}
\caption{Comparison of TEMAD with other mainstream. SA denotes the dataset containing only single-abutment cases, while SMA includes both single- and multi-abutment samples.}
  \label{Diff_model}
\end{table*}

\begin{figure}\centering
\includegraphics[scale=0.35]{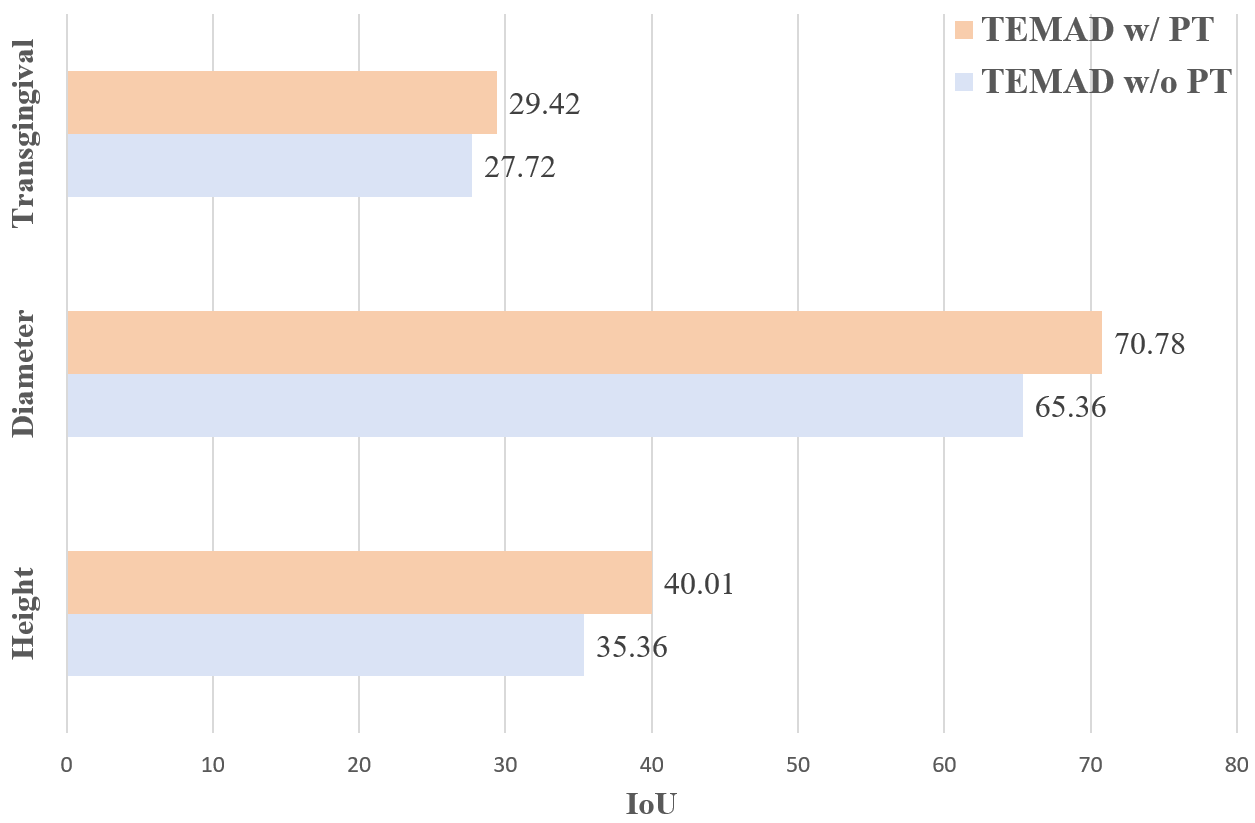}
\caption{Ablation study on encoder pretraining. The horizontal axis denotes IoU performance, while the vertical axis corresponds to the three key abutment parameters, including transgingival, diameter, and height. w/o PT and w/ PT indicate model variants trained without and with encoder pretraining, respectively.}
\label{wo_pretraining}
\end{figure}

\paragraph{\textbf{Ablation Study on Encoder Pretraining}}
We first visualize the masked reconstruction results to examine the effectiveness of the self-supervised encoder pretraining. As shown in Fig~\ref{reconstruction_vis}, using a random masking ratio of 50\%, the pretrained encoder is able to effectively recover the overall geometric structure of the original point cloud, indicating that meaningful global and local anatomical priors are captured during pretraining.
We further evaluate its impact on downstream regression by comparing models trained with pretrained initialization (w/ PT) and from scratch (w/o PT). As shown in Fig~\ref{wo_pretraining}, encoder pretraining consistently improves IoU performance across all three key parameters, with more pronounced gains for diameter and height, both exceeding 5\%. These results demonstrate that geometry-aware initialization leads to more stable optimization and enhanced modeling of subtle anatomical variations in multi-abutment scenarios.

\paragraph{\textbf{Comparison to State-of-the-art Methods.}}
To further evaluate the effectiveness of TEMAD, we conduct comparisons with several state-of-the-art methods, including point cloud-based methods PointNet~\cite{qi2017pointnet}, PointNet++~\cite{qi2017pointnet++}, PointMAE~\cite{pang2022masked}, PointMamba~\cite{liang2024pointmamba}, PointFEMAE~\cite{zha2024towards}, and mesh-based methods MeshMAE~\cite{liang2022meshmae}, TCEAD~\cite{zheng2026text} and SS$A^3$D~\cite{zheng2025ssa3d}. 
We evaluate all methods on both the single-abutment dataset (SA) and the mixed dataset (SMA).
The quantitative results are summarized in Table~\ref{Diff_model}. 

In the SA setting, TEMAD achieves the best performance on the transgingival and diameter parameters, showing clear improvements over previous point-based and mesh-based representation learning methods. In particular, the proposed framework yields consistent gains of approximately 2–6 percentage points on transgingival prediction and over 8 percentage points on diameter compared with the strongest baselines. Althoh slighugtly lower than the best-performing model PointMamba on the height parameter, TEMAD still maintains competitive accuracy while providing more balanced performance across all parameters.
In the more challenging SMA setting, TEMAD demonstrates stronger robustness and generalization ability. The proposed method consistently outperforms all comparison approaches across all three parameters, with performance gains ranging from around 2 to over 20 percentage points. Notably, the improvement is most significant for the height parameter, indicating that TEMAD is more effective in modeling complex inter-site geometric dependencies when multiple abutments are present.
These results highlight the advantage of the proposed framework in learning coordinated structural representations across multiple implant sites, enabling more stable and accurate parameter prediction in complex clinical scenarios.

To further analyze performance in the SMA setting, Fig.~\ref{Multi_Compare} presents parameter-wise IoU comparisons across discrete value intervals. TEMAD consistently outperforms TCEAD for the transgingival parameter, indicating more stable prediction across varying anatomical conditions. For diameter, the advantage is most evident in the clinically common range of 4.0–5.5\,mm. Notably, TEMAD maintains reliable accuracy for larger height values (9.0–12.0\,mm), while TCEAD shows near-zero IoU in these regions, suggesting limited long-range structural modeling capability. Overall, these results demonstrate that TEMAD achieves more balanced and robust parameter prediction in complex multi-abutment scenarios.

\section{Conclusions}
%In this paper, we present TEMAD, a text-conditioned multi-expert regression framework for fully automated multi-abutment design. The proposed approach integrates automatic implant site localization, position-aware feature modulation, and implant system–constrained regression within a unified end-to-end pipeline. By coupling the Implant Site Identification Network (ISIN) with Tooth-Conditioned FiLM (TC-FiLM) and a System-Prompted Mixture-of-Experts (SPMoE) mechanism, TEMAD can design multi-abutment simultaneous from a single intraoral scan while maintaining system compatibility. Experimental results on large-scale single- and multi-abutment datasets demonstrate consistent improvements over existing methods. These findings suggest that the proposed framework provides a practical and scalable solution for automated multi-tooth implant rehabilitation.
In this paper, we present TEMAD, a text-conditioned multi-expert regression framework for fully automated multi-abutment design. The proposed approach consists of two key components: an Implant Site Identification Network (ISIN) for automatic localization of implant sites, and a Multi-Abutment Regression Network (MARN) for simultaneous system-aware parameter prediction. By integrating Tooth-Conditioned FiLM (TC-FiLM) and a System-Prompted Mixture-of-Experts (SPMoE) mechanism, the framework enables coordinated geometric representation learning across multiple implant sites within a unified end-to-end pipeline.
Extensive experiments on both single- and multi-abutment datasets demonstrate that TEMAD achieves consistent performance improvements over existing methods, with particularly strong robustness in complex multi-site scenarios. These results indicate the effectiveness of the proposed design in modeling inter-site geometric dependencies and support its potential as a practical and scalable solution for automated implant rehabilitation planning.

\bibliographystyle{model5-names}
\bibliography{sample-base}

\end{document}